\documentclass[11pt]{article}

\usepackage[final]{acl}

\usepackage{times}
\usepackage{latexsym}
\usepackage[table]{xcolor}
\usepackage{enumitem}

\usepackage[T1]{fontenc}
\usepackage[utf8]{inputenc}
\usepackage{microtype}
\usepackage{inconsolata}
\usepackage{float}

\usepackage{algorithm}
\usepackage{algpseudocode}

\usepackage{booktabs}
\usepackage{array}

\usepackage{graphicx}
\usepackage{rotating}
\usepackage{multirow}
\usepackage{threeparttable}
\usepackage{makecell}

\definecolor{oursgray}{gray}{0.94}
\definecolor{HeadGray}{RGB}{245,245,245}
\definecolor{OursBlue}{RGB}{238,245,255}

\usepackage{amsmath}
\usepackage{amssymb}

\usepackage{xspace}

\graphicspath{{figures/}}

\newcommand{\pms}{\ensuremath{\pm}}

\title{When and What to Teach: Budget-Aware Online Adaptation for Web Agents}

\author{%
  \textbf{Jianwei Zhang\textsuperscript{1,\dag}, Sihan Cao\textsuperscript{1,\dag}, Pengcheng Zheng\textsuperscript{1}, Ya Wen\textsuperscript{1}, Pei Ke\textsuperscript{1}} \\
  \textbf{Kuien Liu\textsuperscript{2}, Shen Gao\textsuperscript{1}, Wei Dong\textsuperscript{3}, Yang Yang\textsuperscript{1}, Chaoning Zhang\textsuperscript{1,*}} \\
  \textsuperscript{1}University of Electronic Science and Technology of China,\\
  \textsuperscript{2}Institute of Software Chinese Academy of Sciences,\\
  \textsuperscript{3}Xi'an University of Architecture and Technology\\
  \texttt{zjw5428c@gmail.com, 2023080903002@std.uestc.edu.cn, }\\
\texttt{chaoningzhang1990@gmail.com}
}

\begin{document}


\maketitle

\begingroup
\renewcommand{\thefootnote}{\fnsymbol{footnote}}
\footnotetext[2]{Equal contribution.}
\footnotetext[1]{Corresponding author.}
\endgroup

\begin{abstract}
Web agents have achieved significant success in automating complex internet tasks but deploying them in real-world environments requires continuous online adaptation. Given that deploying powerful proprietary models remains commercially cost-prohibitive, practitioners must rely on lightweight local models that evolve post-deployment via online teaching from a stronger teacher. However, standard interactive feedback imposes prohibitive costs. We show that conventional trajectory-level preference optimization wastes budget on both unresolvable episodes and redundant execution turns. To resolve these inefficiencies, we propose \textbf{Score-Guided Online Teaching with Budgeted Trajectory Trimming}, a budget-aware framework that systematically orchestrates \textbf{when} and \textbf{what} to teach. Specifically, our framework integrates a solvability-aware teacher gate to dictate \textbf{when} to query the teacher model and a score-guided turn selection mechanism to decide \textbf{what} informative turns to retain. Extensive experiments on MiniWoB and TimeWarp demonstrate that our method achieves comparable first-pass success while reducing teacher calls by 22.6\% and student training compute by 52.1\% on average. Our code is available at \url{https://github.com/zjw131f1fc/budgeted-online-teaching}.
\end{abstract}

\section{Introduction}

Web agents~\cite{wang2026agent,zhang2026lightweight,wei2025webagent,krupp2025quantifying} are autonomous software entities designed to navigate the internet and execute complex digital workflows by perceiving webpage environments and executing concrete browser actions such as clicking, typing, or scrolling. By coupling LLM-based web reasoning with sequential browser interactions, these systems have demonstrated significant promise in automating real-world internet tasks~\cite{nakano2021webgpt, yao2022webshop, yao2023react}. However, static offline training often fails to generalize when agents are deployed in real interactive environments characterized by evolving user interfaces, stochastic network behaviors, and unseen workflows~\cite{zhou2024webarena, deng2023mind2web, xie2024osworld, koh2024visualwebarena}. Meanwhile, directly deploying powerful proprietary models for every task remains commercially cost-prohibitive, so practitioners rely on lightweight local models that must adapt and evolve post-deployment.

A natural solution to mitigate these deployment-phase errors is online teaching. Under this paradigm, whenever the student encounters an execution failure, the teacher is queried to produce a corrective demonstration, thereby updating the student before the next streaming task arrives~\cite{hinton2015distilling}. Unlike standard online reinforcement learning, where the agent explores freely, here the agent is already deployed and first-pass success matters on every episode. This deployment-time online adaptation setting remains, to our knowledge, under-explored for web agents. The most intuitive implementation to realize this workflow is to pair every student failure with a full teacher trajectory and directly feed both elements into a trajectory-level preference optimization objective such as DPO~\cite{rafailov2024direct}.

Nevertheless, we argue that such an unselective, full-trajectory strategy inherently suffers from inefficient supervision allocation, treating expert supervision as an infinite commodity rather than a strictly bottlenecked, strategic resource. This inefficiency manifests acutely at two distinct granularities, corresponding to the critical yet fundamentally overlooked dimensions of \textbf{when} and \textbf{what} to teach. 
At the episode level (\textit{when}), the baseline treats all student failures as equally worth querying, ignoring that the teacher itself is unstable on many tasks and will fail to produce a usable corrective trajectory, wasting the query budget without return~\cite{zhang2016query, hoque2022thriftydagger}. 
At the trajectory level (\textit{what}), it indiscriminately trains on the entire action history, overlooking that long-horizon web sequences are typically riddled with redundant navigation pauses and non-essential actions~\cite{belkhale2023data}. Striking a budget-aware balance across both axes simultaneously remains an open and critical challenge.

To address this issue, we present \textbf{Score-Guided Online Teaching with Budgeted Trajectory Trimming}, a budget-aware online adaptation framework that systematically orchestrates \textbf{when} and \textbf{what} to teach. Our framework integrates a solvability-aware teacher gate to determine \textit{when} to query the teacher model. Tactically, we devise a score-guided turn selection mechanism to decide \textit{what} to teach by retaining the most informative turns on both sides. Crucially, both components are fully compatible with various preference optimization objectives, allowing our framework to be seamlessly applied to standard criteria such as DPO~\cite{rafailov2024direct} and SimPO~\cite{meng2024simpo}. Experiments on MiniWoB~\cite{liu2018reinforcement} and TimeWarp~\cite{ishmam2026timewarp} show that our method achieves comparable online first-pass success to the full-trajectory baselines while reducing teacher calls by 22.6\% and student training compute by 52.1\% on average.

The contributions of this paper are:

\begin{itemize}
\item[$\bullet$] We formulate online web-agent teaching as a budget allocation problem along two axes: \textit{when} to query the teacher and \textit{what} turns to retain for training.

\item[$\bullet$] We introduce a solvability-aware teacher gate that reduces unnecessary teacher queries based on historical teacher outcomes.

\item[$\bullet$] We introduce a score-guided turn selection that reduces redundant training tokens based on the student's current log-probability.

\item[$\bullet$] Experiments on two online web-agent benchmarks validate the framework, showing comparable first-pass success with substantially reduced teacher-query and student-training costs.
\end{itemize}

\section{Related Work}

\subsection{Web Agents and Online Adaptation}
Online adaptation for web agents differs from standard online reinforcement learning, where the agent explores freely to maximize long-term reward. In our setting the agent is already deployed, first-pass success matters on every episode, and teacher supervision carries real cost. Early web-agent work such as WebGPT~\cite{nakano2021webgpt} and WebShop~\cite{yao2022webshop} learns primarily from successful demonstrations. LEAP~\cite{choudhury2025better} introduces iterative fine-tuning from privileged AI feedback on student failures, but assumes unrestricted teacher access and optimizes for final model quality rather than first-pass success on each episode as it arrives. Other approaches improve agents through inference-time experience reuse without gradient updates~\cite{shinn2023reflexion, zhao2024expel, wang2023voyager}. On the training side, DPO~\cite{rafailov2024direct} replaces explicit reward modeling~\cite{christiano2017deep, schulman2017proximal} with direct preference optimization, and recent work extends it to trajectory-level, segment-level, or step-level agent objectives~\cite{shi2024direct, kong2025sdpo, lai2024step, xiong2024watch, song2024trial,cao2026language,zhang2026rcp}. These methods focus on how to construct the preference loss from a given trajectory pool. We instead study which trajectories to collect and which turns to retain under budget constraints.

\subsection{Active and Budget-Aware Imitation}
Selectively requesting expert supervision originates from active learning~\cite{settles2009active} and active imitation learning, where learners query costly expert interventions only on informative states. DAgger~\cite{ross2011reduction} iteratively queries expert actions on learner-visited states to mitigate compounding errors, but it assumes unrestricted expert access. Subsequent variations reduce this cost by gating queries on safety or novelty signals. For instance, SafeDAgger~\cite{zhang2016query} queries only on predicted deviations, ThriftyDAgger~\cite{hoque2022thriftydagger} targets an explicit intervention budget, and HG-DAgger~\cite{kelly2019hg} allows human-gated interventions in embodied control. These methods decide whether to query at the level of individual states using local uncertainty signals. In contrast, our teacher gate estimates whether the current failure is teacher-solvable based on historical outcomes. Furthermore, none of these methods address what to retain once a demonstration is collected. While prior work on demonstration quality~\cite{belkhale2023data,zhang2026immunizing} filters entire episodes, we focus on a finer turn-level selection mechanism.

\subsection{Difficulty-Aware Data Selection for Preference Learning}
A separate line of work uses the current model's own scores as a signal for selecting or weighting training data. Perplexity- and log-probability-based filtering has been used to curate pretraining and fine-tuning corpora by retaining examples the model finds informative~\cite{ankner2025perplexed, xie2023doremi, marion2023less}, and hard-example mining and curriculum strategies have long been used to focus training on instances the current model gets wrong~\cite{shrivastava2016training, bengio2009curriculum}. In preference learning, recent work moves from full-response DPO toward finer-grained credit assignment, including step-level~\cite{lai2024step, xiong2024watch}, segment-level~\cite{kong2025sdpo}, and token-level~\cite{zeng2024token} objectives that re-weight or restrict the contrastive signal to portions of the response. These approaches share the high-level intuition of concentrating supervision on informative portions, but they typically apply a single criterion uniformly and operate at sub-action granularity. Our turn-level selection differs by operating at the interaction-turn level and applying a \emph{dual} criterion: hardest turns on the positive side, most-likely-to-repeat turns on the negative side.
\section{Method}

\subsection{Problem Formulation}

We consider a fixed-order online task stream $\{x_t\}_{t=1}^{T}$~\cite{shalev2025online}, where each task $x_t$ corresponds to a web navigation goal. The student policy $\pi_\theta$ processes tasks sequentially in a first-pass manner: at episode $t$, the student executes independently and produces trajectory $\tau_t^S = (s_1, a_1, \ldots, s_L, a_L)$, where $s_i$ denotes the page state and $a_i$ the student action. If the episode succeeds, the system records the result and continues. If it fails, the system must decide whether to allocate a teacher intervention. The teacher policy $\pi_T$ (e.g., GPT-4o~\cite{hurst2024gpt}), when invoked, executes independently on the same task and produces teacher trajectory $\tau_t^T$. If the teacher succeeds, $(\tau_t^T,\ \tau_t^S)$ forms a matched positive-negative trajectory pair that can be used to update the student. Let $\mathcal{R}_t \in \{0,1\}$ denote whether $\pi_\theta$ succeeds on task $x_t$ at its first attempt. Our objective is to maximize online first-pass success under a teacher-query budget $B_T$ and a student training-token budget $B_S$:
\begin{equation}
  \max_{\theta}\ \sum_{t=1}^{T} \mathcal{R}_t,
  \quad \text{s.t.}\ N_T \leq B_T,\ N_S \leq B_S,
\end{equation}
where $N_T$ is the total number of teacher queries and $N_S$ is the total number of student training tokens.

\subsection{Method Overview}
\begin{figure*}[t]
    \centering
    \includegraphics[width=\textwidth]{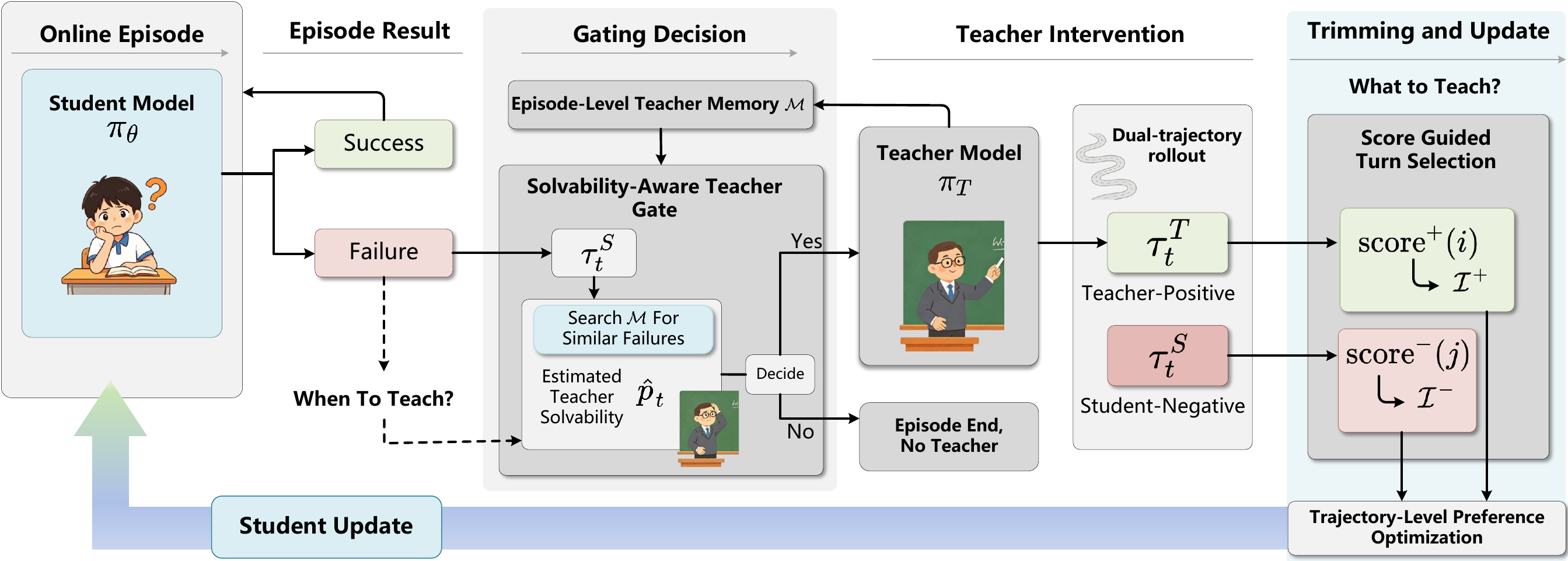}
    \caption{Overview of Score-Guided Online Teaching with Budgeted Trajectory Trimming. The framework first utilizes a \textbf{solvability-aware teacher gate} to decide whether a student failure warrants a teacher query. It then applies a \textbf{score-guided turn selection mechanism} to trim the resulting matched trajectory pair using the student log-probability before performing a trajectory-level preference optimization update.}
    \label{fig:overview}
\end{figure*}

We find that straightforwardly querying the teacher after every student failure and optimizing full trajectories indiscriminately wastes budget at both the episode and turn levels. This inefficiency arises because some failures are inherently unresolvable for the teacher, while even successful rollouts contain redundant, task-specific details with minimal learning value. To resolve these inefficiencies, we organize online teaching into a two-stage mechanism as illustrated in Figure~\ref{fig:overview}. The solvability-aware teacher gate first operates at the episode level to decide whether a student failure warrants an expert rollout. Subsequently, the score-guided turn selection mechanism functions at the turn level to filter the most informative turns from both the positive and negative trajectories, assembling a trimmed preference pair. This optimized pair is then utilized to update the student via a trajectory-level preference optimization objective. Algorithm~\ref{alg:online-step} systematically summarizes the entire end-to-end execution process for a single real-time online step.

\begin{algorithm}[t]
\small
\caption{One online step at task $x_t$.}
\label{alg:online-step}
\begin{algorithmic}[1]
\State Roll out student $\tau_t^S \sim \pi_\theta(x_t)$
\If{$\tau_t^S$ succeeds} \State \Return
\EndIf
\State Compute teacher solvability $\hat{p}_t$ from $\mathcal{M}$
\If{$\mathcal{M}=\emptyset$ \textbf{or} $\sum_{i \in \mathcal{N}_k} w_i < \varepsilon_{\rm exp}$}
  \State \textbf{accept} \Comment{exploration fallback}
\ElsIf{$\hat{p}_t < \lambda$}
  \State \Return \Comment{skip teacher query}
\EndIf
\State Roll out teacher $\tau_t^T \sim \pi_T(x_t)$
\State Append $(\phi(x_t),\ \mathbb{1}[\tau_t^T\text{ succeeds}])$ to $\mathcal{M}$
\If{$\tau_t^T$ succeeds}
  \State Score turns; select $\mathcal{I}^+, \mathcal{I}^-$
  \State Build trimmed pair $(\tilde{\tau}_t^T, \tilde{\tau}_t^S)$
  \State Update $\theta$ via Eq.~\ref{eq:final-loss} or Eq.~\ref{eq:simpo-loss}
\EndIf
\end{algorithmic}
\end{algorithm}

\subsection{Solvability-Aware Teacher Query Gating}

The value of a teacher query depends on whether the current failure can be converted into an effective teaching package: if the teacher is also unstable on the current task, querying it only consumes budget without producing a usable matched positive trajectory. To address this, we maintain an \textbf{episode-level teacher memory} $\mathcal{M}$ that records teacher outcomes on historical failures. Each failure is keyed by a semantic task representation $\phi(x_i) \in \mathbb{R}^d$ that encodes the task name and goal description, and $\mathcal{M}$ stores entries $\{(\phi(x_i),\ y_i)\}$, where $y_i = 1$ if the corresponding teacher rollout produced an accepted matched-positive package and $y_i = 0$ otherwise. When a new failure arrives on task $x_t$, the system retrieves $k$ nearest neighbors~\cite{cover1967nearest} from $\mathcal{M}$ and estimates the \textbf{teacher solvability} of the current failure:
\begin{equation}
  \hat{p}_t = \frac{\sum_{i \in \mathcal{N}_k(x_t)} y_i \cdot w_i}{\sum_{i \in \mathcal{N}_k(x_t)} w_i},
\end{equation}
where $w_i = \exp(-d(\phi(x_t), \phi(x_i))\,/\,\kappa)$ is a similarity weight with semantic distance $d(\cdot,\cdot)$ (cosine distance over a sentence-embedding model) and bandwidth $\kappa$. The gate accepts the failure and issues a teacher query when historical evidence supports teacher solvability, defined by $\hat{p}_t \geq \lambda$, and suppresses it otherwise. To avoid prematurely abandoning under-sampled regions, this criterion is overridden when retrieval evidence is sparse. Specifically, if $\mathcal{M}$ is empty or $\sum_{i \in \mathcal{N}_k} w_i < \varepsilon_{\rm exp}$, the failure is unconditionally accepted. The teacher gate thus concentrates the limited teacher-query budget on failures more likely to yield effective teaching packages, while keeping the memory adaptive to drift in the failure distribution.

\subsection{Score-Guided Turn Selection}
\label{sec:turn-selection}
After the teacher succeeds, the system obtains a teacher-positive trajectory $\tau_t^T = (s_1^T, a_1^T, \ldots, s_m^T, a_m^T)$ and a student-negative trajectory $\tau_t^S = (s_1^S, a_1^S, \ldots, s_n^S, a_n^S)$. Under a fixed budget of at most $M$ turns per side, the goal is to identify the most informative local sub-trajectories from both sequences. On the \textbf{positive side}, we retain the turns that the current student finds most difficult, as lower student log-probabilities indicate higher incremental learning value~\cite{mindermann2022prioritized}. For each turn $i$ in the teacher trajectory, we compute $\mathrm{score}^+(i) = \log \pi_\theta(a_i^T \mid s_i^T)$ and select the $M$ turns with the lowest scores, denoted as $\mathcal{I}^+ = \operatorname{argmin}_{i, |\mathcal{I}^+|=M} \mathrm{score}^+(i)$. On the \textbf{negative side}, we compute $\mathrm{score}^-(j) = \log \pi_\theta(a_j^S \mid s_j^S)$, where a higher score indicates the current policy is more prone to repeating this erroneous behavior. Mirroring the hard-negative selection principle in contrastive learning~\cite{xiong2020approximate}, we retain the $M$ turns with the highest scores, $\mathcal{I}^- = \operatorname{argmax}_{j, |\mathcal{I}^-|=M} \mathrm{score}^-(j)$. Turns indexed by $\mathcal{I}^+$ and $\mathcal{I}^-$ are extracted and concatenated in their original temporal order, yielding trimmed sub-trajectories $\tilde{\tau}_t^T$ and $\tilde{\tau}_t^S$. While negative-side scores are obtained for free from the student rollout, positive-side scoring requires one additional student forward pass. This selection process thus concentrates the training budget on the local segments least familiar to the current student.

\subsection{Trimmed Trajectory-Level Preference Optimization}

Given the trimmed pair $(\tilde{\tau}_t^T, \tilde{\tau}_t^S)$, we treat $\tilde{\tau}_t^T$ as the preferred and $\tilde{\tau}_t^S$ as the dispreferred trajectory, and define trajectory-level log-probability as the mean of per-turn action log-probabilities, $\log \pi_\theta(\tau) = \tfrac{1}{|\tau|} \sum_i \log \pi_\theta(a_i \mid s_i)$. The trimmed pair is compatible with any trajectory-level preference optimization objective. We instantiate two standard algorithmic choices that differ in whether a reference policy is required.
\begin{table*}[!t]
\centering
\small

\setlength{\tabcolsep}{6pt}
\renewcommand{\arraystretch}{1.05}
\caption{Main results on MiniWoB (125-episode stream) and TimeWarp (150-episode stream). \%T = success as a percentage of teacher success. Upd.\ = student updates triggered. Student PFLOPs = training plus positive-side scoring overhead. T.\ Calls = teacher calls. ``--'' = no training cost.}
\begin{tabular}{@{}lccccc@{}}
\toprule
Method & Success & \%T & Upd. & Student PFLOPs & T. Calls \\
\midrule
\multicolumn{6}{@{}l}{\textit{MiniWoB (125 episodes, teacher ceiling 84/125)}} \\
\midrule
Baseline Student           & 30/125 & 35.72\% &  0 & --    & --  \\
Teacher (GPT-4o)           & 84/125 & 100\%  &  0 & --    & --  \\
\midrule
Naive SFT                  & 29\pms1.0/125 & 34.52\pms1.19\% & 53\pms2.0 &  6.22\pms0.49 &  95\pms2.0 \\
Trajectory DPO             & 37.5\pms1.5/125 & 44.65\pms1.79\% & 44.5\pms2.5 & 16.66\pms1.23 &  87\pms1.0 \\
~~+~Query Gate             & 38.0\pms1.0/125 & 45.24\pms1.19\% & 42.0\pms1.0 & 14.37\pms0.45 &  70.5\pms1.5 \\
\rowcolor{gray!20}
~~+~Query Gate + Turn Sel. & 40.5\pms0.5/125 & 48.22\pms0.6\% & 36.5\pms1.5 & 8.45\pms0.28 & 65\pms1.0 \\
Trajectory SimPO          & 44.5\pms1.5/125 & 52.98\pms1.79\% & 37\pms1.0 & 10.85\pms0.36 & 76.5\pms2.5 \\
~~+~Query Gate             & 42\pms1.0/125 & 50.00\pms1.19\% & 31\pms2.0 & 8.86\pms0.68 & 55.5\pms1.5 \\
\rowcolor{gray!20}
~~+~Query Gate + Turn Sel. & 40.5\pms0.5/125 & 48.22\pms0.6\% & 39\pms1.0 & 5.94\pms0.18 & 63.5\pms2.5 \\
\midrule
\multicolumn{6}{@{}l}{\textit{TimeWarp (150 episodes, teacher ceiling 73/150)}} \\
\midrule
Baseline Student           & 14/150 & 19.18\% &  0 & --    & --  \\
Teacher (GPT-4o)           & 73/150 & 100\%  &  0 & --    & --  \\
\midrule
Naive SFT                  & 17\pms2.0/150 & 23.29\pms2.74\% & 54\pms3.0 & 22.83\pms1.68 & 128\pms3.0 \\
Trajectory DPO             & 41\pms1.0/150 & 56.16\pms1.37\% & 39\pms2.0 & 60.26\pms1.15 & 105\pms3.0 \\
~~+~Query Gate             & 39.5\pms1.5/150 & 54.11\pms2.06\% & 36\pms1.0 & 59.04\pms1.51 &  82\pms3.0 \\
\rowcolor{gray!20}
~~+~Query Gate + Turn Sel. & 39\pms2.0/150 & 53.42\pms2.74\% & 25\pms2.0 & 20.42\pms1.09 & 76.5\pms4.5 \\
Trajectory SimPO           & 50\pms3.0/150 & 68.49\pms4.11\% & 29.5\pms1.5 & 37.96\pms4.15 & 97.5\pms1.5 \\
~~+~Query Gate             & 46\pms2.0/150 & 63.01\pms2.74\% & 31.5\pms1.5 & 38.17\pms3.3 & 73\pms3.0 \\
\rowcolor{gray!20}
~~+~Query Gate + Turn Sel. & 46\pms3.0/150 & 63.01\pms4.11\% & 36.5\pms2.5 & 19.86\pms1.4 & 77\pms4.0 \\
\bottomrule
\end{tabular}
\label{tab:main}
\end{table*}
\paragraph{DPO.}
DPO~\cite{rafailov2024direct} computes a log-ratio against a reference policy $\pi_{\mathrm{ref}}$:
\begin{equation}
\begin{aligned}
\mathcal{L}_{\mathrm{DPO}}(\theta)
&= -\log \sigma \Bigg(
    \beta \Big(
      \log \tfrac{\pi_\theta(\tilde{\tau}_t^T)}{\pi_{\mathrm{ref}}(\tilde{\tau}_t^T)} \\
      &\quad - \log \tfrac{\pi_\theta(\tilde{\tau}_t^S)}{\pi_{\mathrm{ref}}(\tilde{\tau}_t^S)}
    \Big)
\Bigg).
\end{aligned}
\label{eq:final-loss}
\end{equation}
$\pi_{\mathrm{ref}}$ is held fixed within a single online update step and reset to the merged student checkpoint at the start of every step, so that each update is performed against the most recent stable student, as in iterative preference optimization and online self-play alignment~\cite{guo2024direct, chen2024self}.

\paragraph{SimPO.}
SimPO~\cite{meng2024simpo} removes the reference policy and uses the length-normalized policy log-likelihood directly, with target reward margin $\gamma$:
\begin{equation}
  \begin{aligned}
\mathcal{L}_{\mathrm{SimPO}}(\theta) &= -\log \sigma \Big( \beta \big( \log \pi_\theta(\tilde{\tau}_t^T) \\
&\quad - \log \pi_\theta(\tilde{\tau}_t^S) \big) - \gamma \Big).
\end{aligned}
\label{eq:simpo-loss}
\end{equation}
We set $\gamma=0.5$ throughout, so the trajectory log-probability defined above (length-normalized) is the only signal driving the contrast.

Both objectives operate on the same trimmed pair produced by Stage 2, and are interchangeable carriers for our framework. The query gate and turn selection are independent of this choice.

\section{Experiments}
\label{sec:experiments}

\subsection{Experimental Setup}

\paragraph{Benchmarks.}
Online teaching is fundamentally a \emph{streaming} setting: the student receives tasks one by one and must adapt as the stream unfolds. We therefore organize each benchmark into a fixed-order online task stream and use that stream as a controlled evaluation protocol shared across all methods.
\textbf{MiniWoB}~\citep{liu2018reinforcement} is instantiated as a 125-episode stream covering form filling, email operations, search, and other common web interaction patterns.
\textbf{TimeWarp}~\citep{ishmam2026timewarp} is instantiated as a 150-episode stream with longer and more heterogeneous tasks, placing greater demands on cross-task generalization.
The task ordering is completely fixed offline and reused for every method, so that any difference in observed performance reflects the online learning behavior rather than variation in the task sequence, and all methods report first-pass results on the same evaluation stream of tasks.

\paragraph{Baselines.}
\textbf{Baseline Student} applies no online teaching and reflects the student's initial capability.
\textbf{Teacher Baseline} runs the teacher independently on the same task stream, providing an upper-bound reference.
Among online teaching methods, \textbf{Naive SFT} fine-tunes on matched teacher-positive trajectories without using student failure trajectories as negative signal.
We then consider two trajectory-level preference optimization carriers, applied directly between complete teacher-positive and student-failed trajectories with no selection mechanism: \textbf{Trajectory DPO} and \textbf{Trajectory SimPO}.
On top of each carrier, \textbf{+~Query Gate} adds episode-level teacher query gating, and \textbf{+~Query Gate + Turn Selection} (our full method) further adds score-guided turn-level trimming.

\paragraph{Metrics.}
The primary metric is \textbf{online first-pass success} (number of episodes succeeded on the first attempt). We also report \textbf{\%T} (success as a percentage of teacher success) and \textbf{Upd.} (number of student updates triggered during the stream).
Budget metrics include teacher calls, failed positive resolutions (teacher calls that fail to produce an acceptable matched-positive trajectory), and \textbf{Student Total PFLOPs} (training plus positive-side scoring overhead). PFLOPs are computed under a dense-equivalent assumption from replay-counted tokens. Full accounting is in Appendix~\ref{app:pflops}.

\paragraph{Implementation Details.}
The student model is Qwen2.5-3B-Instruct~\cite{qwen2025qwen25technicalreport} and the teacher is GPT-4o~\cite{hurst2024gpt}.
All rollouts use the standard \texttt{GenericAgent} from AgentLab on top of the BrowserGym ecosystem~\cite{chezelles2025browsergym}. Teacher and student share the same observation and action configuration and differ only in the chat-model backend (OpenAI API vs.\ a local vLLM~\cite{kwon2023efficient} endpoint).
All methods use LoRA~\cite{hu2022lora, dettmers2023qlora} fine-tuning. DPO uses Eq.~\ref{eq:final-loss} with $\beta = 0.1$, and SimPO uses Eq.~\ref{eq:simpo-loss} with $\beta = 2.0$ and $\gamma = 0.5$.
Gate thresholds $\lambda$ / $\varepsilon_{\rm exp}$ are $0.35$ / $0.50$ on MiniWoB and $0.30$ / $0.50$ on TimeWarp.
Turn budgets are top-$3$/side for MiniWoB and top-$4$/side for TimeWarp.
Full hyperparameters, optimizer settings, and agent and gate configuration details are detailed in Appendix~\ref{app:agent-config} and Appendix~\ref{app:hparams}.

\subsection{Main Results}

Table~\ref{tab:main} reports the main results on both benchmarks. Naive SFT attains $29.0\pms1.0$/125 success on MiniWoB and $17.0\pms2.0$/150 on TimeWarp while making $95.0\pms2.0$ and $128.0\pms3.0$ teacher calls, respectively. Preference optimization uses the available supervision more effectively: full-trajectory DPO reaches $37.5\pms1.5$/125 and $41.0\pms1.0$/150, while full-trajectory SimPO reaches $44.5\pms1.5$/125 and $50.0\pms3.0$/150.

On the DPO carrier, adding both the query gate and turn selection reduces MiniWoB Student PFLOPs from $16.66\pms1.23$ to $8.45\pms0.28$ and teacher calls from $87.0\pms1.0$ to $65.0\pms1.0$, while success changes from $37.5\pms1.5$ to $40.5\pms0.5$. On TimeWarp, Student PFLOPs decrease from $60.26\pms1.15$ to $20.42\pms1.09$ and teacher calls from $105.0\pms3.0$ to $76.5\pms4.5$, while success changes from $41.0\pms1.0$ to $39.0\pms2.0$.

The same efficiency pattern holds with SimPO. On MiniWoB, the full method reduces Student PFLOPs from $10.85\pms0.36$ to $5.94\pms0.18$ and teacher calls from $76.5\pms2.5$ to $63.5\pms2.5$, with success changing from $44.5\pms1.5$ to $40.5\pms0.5$. On TimeWarp, it reduces Student PFLOPs from $37.96\pms4.15$ to $19.86\pms1.40$ and teacher calls from $97.5\pms1.5$ to $77.0\pms4.0$, with success changing from $50.0\pms3.0$ to $46.0\pms3.0$. Across these four same-carrier comparisons, teacher calls decrease by 22.6\% and student compute by 52.1\% on average.

The MiniWoB teacher-side breakdown in Table~\ref{tab:miniwob-teacher} further shows that failed positive resolutions decrease from 41 without gating to 29 with the query gate and 28 with both mechanisms.

\begin{table}[!t]
\centering
\small
\setlength{\tabcolsep}{4pt}
\renewcommand{\arraystretch}{1.05}
\caption{MiniWoB teacher-side statistics on the DPO carrier. T.\ Calls = teacher calls, Failed Res.\ = teacher calls that did not produce an acceptable matched-positive trajectory, Upd.\ = number of student updates triggered.}
\begin{tabular}{@{}lccc@{}}
\toprule
Method & \makecell[c]{T. Calls} & \makecell[c]{Failed Res.} & \makecell[c]{Upd.} \\
\midrule
Trajectory DPO            & 87 & 41 & 46 \\
~~+~Query Gate            & 70 & 29 & 41 \\
~~+~Query Gate + Turn Sel.& 66 & 28 & 38 \\
\bottomrule
\end{tabular}
\label{tab:miniwob-teacher}
\end{table}

\subsection{Ablation Studies}

\subsubsection{Query Efficiency of the Teacher Gate}

To isolate the gate's allocation behavior from subsequent student updates, we replay the fixed failure stream from the completed TimeWarp DPO+Gate run. The stream contains 111 replayable student failures, of which 40 have successful teacher-positive oracles. We compare the number of teacher queries required by the solvability-aware gate and by random ordering, averaged over 128 random seeds, to reach the same cumulative number of oracle-positive hits.
\begin{figure}[t]
    \centering
    \includegraphics[width=0.9\linewidth]{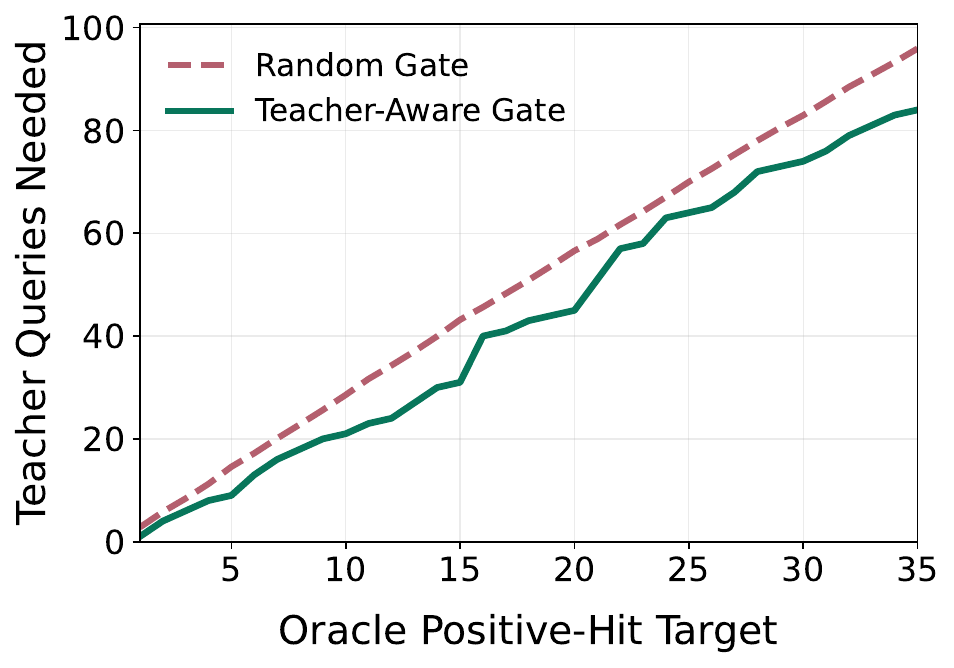}
    \caption{Teacher queries required to reach each oracle-positive hit target on the TimeWarp offline replay stream. The solvability-aware teacher gate uses fewer queries than random ordering at every reachable target; the random curve is averaged over 128 seeds.}
    \label{fig:gate}
\end{figure}
As shown in Fig.~\ref{fig:gate}, the solvability-aware gate requires fewer queries than random ordering at all 35 hit targets reached by the default gate. For example, it reaches 20 oracle-positive hits in 45 queries, compared with 56.6 queries for random ordering, a 20.5\% reduction. At the highest common target of 35 hits, it requires 84 queries compared with 95.9 for random ordering, a 12.4\% reduction. This indicates that the gate preferentially allocates calls to failures for which usable teacher supervision is more likely to be available.

\subsubsection{Turn Selection Ablation}

This ablation tests whether trimming each side of the trajectory pair contributes to efficiency.
All variants include the teacher-aware query gate and use the DPO carrier on MiniWoB (top-3/side).
We progressively enable score-guided turn selection on each side:
\textbf{No Trim} keeps the full trajectory on both sides (gate-only reference),
\textbf{Trim Neg Only} applies selection on the negative side while keeping the full positive trajectory,
\textbf{Trim Pos Only} applies selection on the positive side while keeping the full negative trajectory,
\textbf{Trim Both} (full method) trims both sides.

Table~\ref{tab:ablation-trim} shows the results.
Negative-side trimming is the dominant source of efficiency on MiniWoB: trimming only the negative side reduces student PFLOPs by 36\% while success changes from 38/125 to 40/125.
Positive-side trimming contributes a smaller reduction because MiniWoB teacher trajectories are already short (avg ${\sim}3$ turns), leaving little room for the top-3 budget to discard.
Trimming both sides yields the largest overall saving ($-$41\% PFLOPs).

\begin{table}[t]
\centering
\small
\setlength{\tabcolsep}{4pt}
\renewcommand{\arraystretch}{1.05}
\caption{Turn selection ablation on MiniWoB (DPO carrier, top-3/side). All variants include the query gate. Student PFLOPs = Student Total PFLOPs.}
\begin{tabular}{@{}lcccc@{}}
\toprule
Method & Success & \makecell[c]{Student\\ PFLOPs} & \makecell[c]{$\Delta$ PFLOPs} & \makecell[c]{T. Calls} \\
\midrule
No Trim             & 38/125 & 14.37 & --      & 70 \\
Trim Pos Only       & 37/125 & 13.25 & $-$8\%  & 67 \\
Trim Neg Only       & 40/125 & 9.13  & $-$36\% & 65 \\
Trim Both           & 40/125 & 8.45 & $-$41\% & 65 \\
\bottomrule
\end{tabular}
\label{tab:ablation-trim}
\end{table}

\subsubsection{Sensitivity to Gate Threshold}

The query-efficiency ablation shows that the solvability-aware teacher gate allocates a fixed query budget more effectively than random selection, but this conclusion could depend on a single threshold operating point. To test robustness on TimeWarp, we replay the fixed failure stream from the completed DPO+Gate run, which contains 111 replayable student failures and 40 successful teacher-positive oracles. We fix $k=10$ and $\kappa=0.16$ and sweep $\lambda$ and $\varepsilon_{\rm exp}$ independently around the default setting. For each operating point, the random baseline uniformly selects the same number of failures as the gate and reports expected oracle hits over 128 seeds.

Table~\ref{tab:ablation-gate-thresh-timewarp} reports five nearby threshold combinations. The first three rows fix $\varepsilon_{\rm exp}=0.50$ and vary $\lambda$; the last two fix $\lambda=0.30$ and vary $\varepsilon_{\rm exp}$. At every operating point, the solvability-aware gate obtains more oracle hits than random selection at the same query budget, with an advantage from $+2.5$ to $+5.9$ hits. The default TimeWarp setting $(\lambda,\varepsilon_{\rm exp})=(0.30,0.50)$ exactly reproduces the online gate record with 85 queries and 35 oracle hits, compared with 31.1 expected hits under random selection. Small perturbations shift the budget--coverage tradeoff but do not reverse the direction of the advantage.

\begin{table}[t]
\centering
\small
\setlength{\tabcolsep}{4pt}
\renewcommand{\arraystretch}{1.05}
\caption{Gate threshold sensitivity on the TimeWarp offline replay stream constructed from the completed DPO+Gate run. Random Hits is the expected number of teacher-positive oracle hits for a random gate at the same query budget, averaged over 128 seeds. $\dagger$ denotes the default TimeWarp setting.}
\begin{tabular}{@{}lcccc@{}}
\toprule
$(\lambda,\ \varepsilon_{\rm exp})$ & Queries & Gate Hits & Random Hits & $\Delta$ Hits \\
\midrule
(0.25, 0.50) & 92 & 36 & 33.5 & +2.5 \\
(0.30, 0.50)$^\dagger$ & 85 & 35 & 31.1 & +3.9 \\
(0.35, 0.50) & 79 & 35 & 29.1 & +5.9 \\
(0.30, 0.40) & 83 & 35 & 30.6 & +4.4 \\
(0.30, 0.60) & 88 & 36 & 32.1 & +3.9 \\
\bottomrule
\end{tabular}
\label{tab:ablation-gate-thresh-timewarp}
\end{table}

\subsubsection{Sensitivity to Turn Budget}

The MiniWoB main experiment uses top-3/side and TimeWarp uses top-4/side.
To verify these are not isolated operating points, Table~\ref{tab:ablation-budget} reports MiniWoB results across turn budgets on the DPO carrier.
All three trimmed configurations match or exceed the untrimmed full-trajectory reference at substantially lower student PFLOPs, indicating that the benefit of turn-level trimming is robust across budget settings.
Note that different turn budgets affect the student's subsequent policy evolution and thus the failure distribution and total training volume, so success and student total PFLOPs need not vary monotonically with budget.

\begin{table}[t]
\centering
\small
\setlength{\tabcolsep}{4pt}
\caption{Turn budget sensitivity on MiniWoB (DPO carrier). Student PFLOPs = Student Total PFLOPs.}

\begin{tabular}{@{}lccc@{}}
\toprule
Method & Success & \makecell[c]{Student PFLOPs} & \makecell[c]{T. Calls} \\
\midrule
top-2/side      & 38/125 &  5.17 & 63 \\
top-3/side      & 40/125 & 8.45 & 65 \\
top-4/side      & 41/125 & 9.84 & 61 \\
Full Trajectory & 38/125 & 14.37 & 70 \\
\bottomrule
\end{tabular}
\label{tab:ablation-budget}
\end{table}

\subsubsection{Sensitivity to the Preference-Optimization Coefficient}

Both DPO and SimPO have a coefficient $\beta$ that scales the preference logit, though the two operate on different scales (Section~\ref{sec:experiments}, Implementation Details). To check whether our findings are sensitive to this hyperparameter, we sweep $\beta$ on MiniWoB with the full method, using DPO as the carrier. Table~\ref{tab:ablation-beta} reports the result. Across the entire $0.05$--$0.5$ range, success and student PFLOPs both stay close to the default operating point, with $\beta=0.5$ reaching the highest success at $42/125$ and the default $\beta=0.1$ at $40/125$. The conclusion that score-guided turn selection improves online first-pass performance under reduced budget is therefore not sensitive to the exact $\beta$ choice.

\begin{table}[t]
\centering
\small
\setlength{\tabcolsep}{4pt}
\caption{Sensitivity to the preference-optimization coefficient $\beta$ on MiniWoB (full method, DPO carrier, top-3/side). $\dagger$ = default setting. Student PFLOPs = Student Total PFLOPs.}

\begin{tabular}{@{}lccc@{}}
\toprule
$\beta$ & Success & \makecell[c]{Student PFLOPs} & \makecell[c]{T. Calls} \\
\midrule
0.05      & 39/125 & 8.70 & 65 \\
0.1$^\dagger$  & 40/125 & 8.45 & 65 \\
0.2      & 39/125 &  8.35 & 63 \\
0.5      & 42/125 & 9.23 & 64 \\
\bottomrule
\end{tabular}
\label{tab:ablation-beta}
\end{table}

\subsection{Qualitative Analysis}
\label{sec:qualitative}

Inspecting individual teaching packages clarifies what score-guided turn selection actually keeps and discards, and the retained turns consistently track the student's current weakness across both benchmarks, which a positional heuristic such as keeping the most recent turns cannot reproduce. Full trimmed pairs with per-turn scores are given in Appendix~\ref{app:qualitative-full}.

A representative MiniWoB case is \texttt{use-autocomplete}, where the goal is to enter an item beginning with ``Gua'' and ending with ``uam''. The teacher types the prefix to trigger the autocomplete dropdown, picks the suggestion \texttt{Guam}, edits it into \texttt{Guaguam}, and submits. Score-guided selection retains the autocomplete-trigger and the subsequent text edit while dropping the trailing submit, which the student already handles. The matching student-failure trajectory submits the prefix prematurely, retries, and eventually enters the malformed word \texttt{Guamau}. The selector retains exactly these turns, capturing the failure mode the student is most likely to repeat. The trimmed pair thus concentrates the imitation signal on the corrective subroutine and the contrastive signal on the recurring error.

The same behavior carries over to longer TimeWarp tasks. On a multi-hop search task that requires opening the fourth result for ``Utah'', identifying ``New Orleans'' as the first city in the article, searching for it, and reporting the final state name from its fourth result, the teacher's complete trajectory contains routine search-box fills and Go clicks. Selection drops the initial search-box fill and both routine \texttt{Go} clicks, while retaining the two cross-article jumps, the intermediate \texttt{New Orleans} query, and the final answer, which together form the actual reasoning chain the student must learn. The matching student trajectory never reaches this chain: it diverts into the Sci/Tech section and opens an unrelated technology article, and the negative side preserves exactly this divergence while dropping the closing ``incomplete task'' message.



\section{Conclusion}
In this work, we investigated online teaching for web agents under realistic budget constraints. Driven by the fact that deploying proprietary models is commercially cost-prohibitive, we enabled lightweight local models to effectively adapt and evolve post-deployment. To eliminate the inefficiencies of unresolvable queries and redundant execution turns in full-trajectory optimization, we realized a two-stage budget allocation framework. Specifically, our method integrates a solvability-aware teacher gate using a k-NN historical memory to filter queries alongside a score-guided turn selection mechanism to retain only highly informative execution segments. This joint formulation allows the student model to systematically establish a clear understanding of when to seek external assistance and what critical behaviors to replicate. Experiments on MiniWoB and TimeWarp show that our framework achieves comparable or superior online success compared to trajectory-level DPO while reducing teacher calls by 22.6\% and student training compute by 52.1\% on average. These substantial gains establish a highly cost-effective paradigm for future scalable online intelligence.

\section*{Limitations}

Our evaluation focuses on simulated web-agent benchmarks. Extending the framework to live websites, multi-modal computer-use tasks, or non-English interfaces is a natural next step but does not require architectural changes.
The teacher gate uses a standard text-embedding model for $k$-NN retrieval; while this worked well in our setting, environments with highly heterogeneous page structures may benefit from domain-adapted embeddings.
The current turn selection criterion is based solely on the student's log-probability. Incorporating additional signals such as state novelty or action diversity could further refine the selection, though our ablations suggest the current criterion already captures the dominant source of variation.
Finally, our framework assumes a single fixed teacher model. Exploring multi-teacher or self-improving teacher configurations is an interesting direction for future work.

\section*{Acknowledgments}

This work was partially supported by the National Natural Science Foundation of China under Grants 62572104 and 62220106008.

\bibliography{custom}

\clearpage
\appendix
\section{Agent and Prompt Configuration}
\label{app:agent-config}

All rollouts use the standard \texttt{GenericAgent} class from AgentLab on top of BrowserGym~\cite{chezelles2025browsergym} with the default \texttt{FLAGS\_GPT\_4o} preset. We did not modify the AgentLab prompt template or action parser. The contributions of this paper are in the teacher query gate, turn scoring and selection, and the student update logic, not in hand-engineered web-agent prompts. Teacher and student agents share the same observation, action, and prompt configuration and differ only in the chat-model backend: the teacher uses the OpenAI API (GPT-4o), and the student uses a local vLLM endpoint serving Qwen2.5-3B-Instruct, with rollout log-probability access exposed by the inference wrapper.

\paragraph{Observation.}
The agent observes the page through the AXTree representation provided by BrowserGym, augmented with the currently focused element, the recent interaction history, the recent action history, and current error logs. Visible and clickable tags are preserved in the AXTree. HTML dumps, screenshots, set-of-mark annotations, coordinate extraction, and historical thoughts are disabled. Concretely, the relevant \texttt{FLAGS\_GPT\_4o.obs} fields are set as: \texttt{use\_ax\_tree=True}, \texttt{use\_html=False}, \texttt{use\_screenshot=False}, \texttt{use\_som=False}, \texttt{use\_focused\_element=True}, \texttt{use\_history=True}, \texttt{use\_action\_history=True}, \texttt{use\_think\_history=False}, \texttt{use\_error\_logs=True}, \texttt{extract\_visible\_tag=True}, \texttt{extract\_clickable\_tag=True}, \texttt{extract\_coords=False}.

\paragraph{Action Space.}
We use the standard bid-based high-level action set from BrowserGym (\texttt{action\_set.subsets=("bid",)}), restricted to single-action steps (\texttt{multiaction=False}). The exact set of available actions depends on the environment: MiniWoB tasks typically expose \texttt{click}, \texttt{fill}, and \texttt{keyboard\_press}. TimeWarp additionally exposes navigation actions such as \texttt{go\_back}, \texttt{goto}, \texttt{tab\_focus}, and \texttt{send\_msg\_to\_user}. Action descriptions are kept in their default short form (\texttt{long\_description=False}, \texttt{individual\_examples=False}).

\paragraph{Prompt Switches.}
We use the default \texttt{FLAGS\_GPT\_4o} prompt switches: \texttt{use\_thinking=True}, \texttt{use\_abstract\_example=True}, \texttt{use\_concrete\_example=True}, \texttt{use\_hints=True}, \texttt{use\_plan=False}, \texttt{use\_memory=False}, \texttt{use\_criticise=False}, \texttt{enable\_chat=False}. The prompt token budget is \texttt{max\_prompt\_tokens=40000} with up to \texttt{max\_trunc\_itr=20} truncation iterations when the budget is exceeded.

\section{Full Hyperparameter and Gate Configuration}
\label{app:hparams}

This appendix lists training, optimizer, and gate configuration details that are condensed in Section~\ref{sec:experiments} (Implementation details).

\paragraph{LoRA and Training Objectives.}
All methods fine-tune the student with LoRA (rank $=16$, scaling $=32$, dropout $=0.05$) on \texttt{q\_proj}, \texttt{k\_proj}, \texttt{v\_proj}, \texttt{o\_proj}, \texttt{gate\_proj}, \texttt{up\_proj}, and \texttt{down\_proj}. Naive SFT applies only an SFT update to an available positive teaching trajectory. DPO uses the standard objective in Eq.~\ref{eq:final-loss} with $\beta=0.1$ and an explicit reference model; the reference is the current stable student state for each update, with the clean base model used for the first update. SimPO uses the reference-free objective in Eq.~\ref{eq:simpo-loss} with $\beta=2.0$ and $\gamma=0.5$. DPO and SimPO use no SFT preheat, auxiliary SFT loss, or anchor SFT examples. After every successful update, the promoted student state is used for the next episode.

\paragraph{Optimizer.}
We use AdamW with learning rate $5 \times 10^{-5}$ on MiniWoB and $1 \times 10^{-4}$ on TimeWarp, warmup ratio $0.05$, effective batch size $8$, and $1$ epoch per online update. Maximum sequence lengths are 4{,}000 tokens for MiniWoB and 7{,}000 tokens for TimeWarp.

\paragraph{Teacher Gate.}
The teacher memory retrieves $k=10$ neighbors with similarity bandwidth $\kappa=0.16$. The solvability threshold $\lambda$ and exploration evidence threshold $\varepsilon_{\rm exp}$ are $0.35$ / $0.50$ on MiniWoB and $0.30$ / $0.50$ on TimeWarp. The semantic task representation $\phi(x)$ is computed by encoding the string \texttt{task=<task\_name>;\ goal=<task\_goal>} with OpenAI \texttt{text-embedding-3-small}~\cite{openai2024embeddings}. The embedding API is called once per gate evaluation, and its token cost is negligible compared to the teacher rollout cost and is not counted in the teacher token totals.

\paragraph{Turn Selection.}
Turn budgets are top-$3$/side for MiniWoB and top-$4$/side for TimeWarp. Negative-side scoring reuses the log-probabilities recorded during the student's own rollout and incurs no additional cost. Positive-side scoring requires one student forward pass per teacher-positive turn.

\section{Full Trimmed Trajectory Pairs}
\label{app:qualitative-full}

This appendix complements the qualitative analysis in Section~\ref{sec:qualitative} by showing the full teacher-positive and student-failure trajectories of the two examples, together with the turns selected by score-guided turn selection on each side. The reported \emph{selection score} is the per-side criterion from Section~\ref{sec:turn-selection}, normalized to $[0,1]$ within the trajectory: on the positive side it ranks turns by lowest student log-probability (turns the student finds most difficult), and on the negative side it ranks turns by highest student log-probability (failure turns the student is most likely to repeat). Selected turns are the top-$M$ turns by score on each side, where $M=3$ on MiniWoB and $M=4$ on TimeWarp, matching the main experimental setup.

\subsection{Example A: \texttt{use-autocomplete-nodelay} (MiniWoB)}

The task instruction is \emph{``Enter an item that starts with `Gua' and ends with `uam'.''} The page exposes a \texttt{Tags} input field, an autocomplete menu that opens on input, and a \texttt{Submit} button. Element bids \texttt{17}, \texttt{18}, and \texttt{24} correspond to the input box, the submit button, and the \texttt{Guam} candidate in the autocomplete menu, respectively.

\begin{table}[h]
\centering
\small
\setlength{\tabcolsep}{4pt}
\renewcommand{\arraystretch}{1.05}
\caption{Teacher-positive trajectory for \texttt{use-autocomplete-nodelay}. Selection retains the autocomplete-trigger and the subsequent text edit, dropping the trailing submit click that the student already handles correctly.}
\begin{tabular}{@{}clcc@{}}
\toprule
\# & Action & Score & Sel. \\
\midrule
0 & \texttt{fill('17', 'Gua', auto=T)}        & 0.702 & \checkmark \\
1 & \texttt{click('24')} (\texttt{Guam})       & 0.065 & \checkmark \\
2 & \texttt{fill('17', 'Guaguam')}             & 1.000 & \checkmark \\
3 & \texttt{click('18')} (\texttt{Submit})     & 0.000 & --         \\
\bottomrule
\end{tabular}
\label{tab:qual-autocomplete-pos}
\end{table}

\begin{table}[h]
\centering
\small
\setlength{\tabcolsep}{4pt}
\renewcommand{\arraystretch}{1.05}
\caption{Student-failure trajectory for \texttt{use-autocomplete-nodelay}. Selection retains two premature submits and the malformed final string \texttt{Guamau}, which together capture the failure modes the student is most likely to repeat.}
\begin{tabular}{@{}clcc@{}}
\toprule
\# & Action & Score & Sel. \\
\midrule
0 & \texttt{fill('17', 'Gua')}                 & 0.389 & --         \\
1 & \texttt{click('18')} (premature submit)    & 0.868 & \checkmark \\
2 & \texttt{fill('17', '')}                    & 0.578 & --         \\
3 & \texttt{fill('17', 'Guat')}                & 0.405 & --         \\
4 & \texttt{click('18')} (premature submit)    & 1.000 & \checkmark \\
5 & \texttt{fill('17', 'Guama')}               & 0.176 & --         \\
6 & \texttt{fill('17', 'Guamau')}              & 0.756 & \checkmark \\
7 & \texttt{fill('17', '')}                    & 0.626 & --         \\
8 & \texttt{fill('17', 'Guamau')}              & 0.440 & --         \\
9 & \texttt{noop()}                            & 0.000 & --         \\
\bottomrule
\end{tabular}
\label{tab:qual-autocomplete-neg}
\end{table}

\subsection{Example B: Multi-hop Search (TimeWarp)}

The task instruction is \emph{``Search `Utah', visit the fourth search result, identify the first city in the article body, search that city, visit the fourth search result, and report the last state mentioned in the visited article.''} The intended chain is: search \texttt{Utah} $\rightarrow$ open the fourth result \texttt{U.S.\ state of Utah begins to accept evacuees from Hurricane Katrina} $\rightarrow$ identify \texttt{New Orleans} as the first city $\rightarrow$ search \texttt{New Orleans} $\rightarrow$ open the fourth result \texttt{Total evacuation of New Orleans planned} $\rightarrow$ report \texttt{Texas} as the last state mentioned.

\begin{table}[h]
\centering
\small
\setlength{\tabcolsep}{4pt}
\renewcommand{\arraystretch}{1.05}
\caption{Teacher-positive trajectory for the multi-hop search task. Selection drops the initial search-box fill and both routine \texttt{Go} clicks, while retaining the two cross-article jumps, the intermediate \texttt{New Orleans} query, and the final answer, which together form the actual reasoning chain of the task.}
\begin{tabular}{@{}p{0.4cm}p{4.5cm}cc@{}}
\toprule
\# & Action & Score & Sel. \\
\midrule
0 & \texttt{fill('257', 'Utah')}                                & 0.000 & --         \\
1 & \texttt{click('258')} (\texttt{Go})                          & 0.034 & --         \\
2 & \texttt{click('132')} open Utah result                       & 0.047 & \checkmark \\
3 & \texttt{fill('151', 'New Orleans')}                          & 0.395 & \checkmark \\
4 & \texttt{click('152')} (\texttt{Go})                          & 0.000 & --         \\
5 & \texttt{click('132')} open New Orleans result                & 0.931 & \checkmark \\
6 & \texttt{send\_msg\_to\_user("Texas")}                        & 1.000 & \checkmark \\
\bottomrule
\end{tabular}
\label{tab:qual-multihop-pos}
\end{table}

\begin{table}[h]
\centering
\small
\setlength{\tabcolsep}{4pt}
\renewcommand{\arraystretch}{1.05}
\caption{Student-failure trajectory for the multi-hop search task. Selection retains the diverging navigation into \texttt{Sci/Tech}, the unrelated tech article, and the misplaced \texttt{New Orleans} search, while dropping the closing ``incomplete task'' message.}
\begin{tabular}{@{}p{0.4cm}p{4.5cm}cc@{}}
\toprule
\# & Action & Score & Sel. \\
\midrule
0 & \texttt{click('48')} enter Sci/Tech                          & 0.515 & \checkmark \\
1 & \texttt{click('48')} re-enter Sci/Tech                       & 1.000 & \checkmark \\
2 & \texttt{click('105')} unrelated tech article                 & 0.453 & \checkmark \\
3 & \texttt{fill('150', 'New Orleans')} in wrong context         & 0.134 & \checkmark \\
4 & \texttt{send\_msg\_to\_user("Incomplete task: ...")}         & 0.000 & --         \\
\bottomrule
\end{tabular}
\label{tab:qual-multihop-neg}
\end{table}

Across both examples the criterion is asymmetric across sides: positive-side selection extracts the corrective subroutine that the student does not yet master, while negative-side selection isolates the divergence pattern the student is most likely to repeat. A positional heuristic such as keeping the most recent turns cannot reproduce either pattern, since the informative segments are interior rather than terminal in both trajectories.

\section{PFLOPs Accounting}
\label{app:pflops}

This appendix details how the \textbf{Student Total PFLOPs} column reported in Section~\ref{sec:experiments} is computed. All numbers are derived from a single accounting pass over the exported per-update datasets, rather than from in-flight runtime estimates.

\paragraph{Dense-equivalent assumption.}
We compute student-side PFLOPs using a dense-transformer equivalent with \(P = 3\times 10^{9}\) parameters, matching the parameter count of Qwen2.5-3B-Instruct. LoRA reduces the number of trainable parameters but the forward and backward passes still involve the full dense weights, so the dense-equivalent count is the appropriate coefficient for compute reporting.

\paragraph{Training PFLOPs.}
Training cost is recomputed from the exported per-update datasets, not from runtime estimates. For each online update, we re-tokenize the dataset files with the student tokenizer under the run's maximum sequence length. We apply method-specific coefficients reflecting the number of forward and backward passes per training token:
\begin{itemize}
\item \textbf{Naive SFT and SimPO}: coefficient $6$ (one forward + one backward pass),
\begin{equation}
\text{Train PFLOPs} = \frac{6 \cdot P \cdot N_{\mathrm{train}}}{10^{15}}.
\end{equation}
\item \textbf{DPO-based methods}: coefficient $8$, because DPO requires two forward passes (policy and reference) plus one backward pass per training token,
\begin{equation}
\text{Train PFLOPs} = \frac{8 \cdot P \cdot N_{\mathrm{train}}}{10^{15}}.
\end{equation}
\end{itemize}
For methods with score-guided turn selection, the positive-side scoring forward pass (coefficient $2$) is added to the objective-dependent training cost:
\begin{equation}
\text{Student Total PFLOPs} = \frac{c_{\mathrm{obj}} \cdot P \cdot N_{\mathrm{train}} + 2 \cdot P \cdot N^{+}_{\mathrm{score}}}{10^{15}},
\end{equation}
where $c_{\mathrm{obj}}=6$ for Naive SFT and SimPO and $c_{\mathrm{obj}}=8$ for DPO. Here, $N^{+}_{\mathrm{score}}$ is the token count of the positive-side scoring requests, read from the per-update score request log. For methods without turn selection, Student Total PFLOPs equals Train PFLOPs directly.

\paragraph{Per-update Tokens.}
Each successful online update trains only on the teaching package produced for the current episode. Previously collected teaching packages are not replayed by the main training path. The accounting pass re-tokenizes each exported per-update dataset and counts its tokens once, on the update where they contribute to the gradient.

\paragraph{Negative-side Scoring is Excluded.}
Negative-side turns are generated by the student policy itself during rollout, so the per-turn log-probabilities \(\log \pi_\theta(a_j^S \mid s_j^S)\) used by the negative-side selector are available as rollout-time inference outputs and do not require a separate forward pass.



\section{Compute, Artifacts, and Responsible NLP}
\label{app:compute-and-artifacts}

\paragraph{Compute Budget.}
All experiments were run on NVIDIA GeForce RTX 4090 GPUs. Each online run colocated student rollout through vLLM and LoRA fine-tuning on one GPU; for DPO, the explicit reference model was held on CPU. Multiple independent runs could execute concurrently on different GPUs. We report artifact-derived Student PFLOPs rather than an aggregate GPU-hour estimate.

\paragraph{Artifacts and Licenses.}
We use the following artifacts strictly within their intended research-use terms:
\begin{itemize}
\item \textbf{Qwen2.5-3B-Instruct}~\cite{qwen2025qwen25technicalreport} (student model), released by Alibaba under the Qwen license, which permits research use.
\item \textbf{GPT-4o}~\cite{hurst2024gpt} (teacher model), accessed through the OpenAI API under the OpenAI Terms of Service.
\item \textbf{OpenAI \texttt{text-embedding-3-small}}~\cite{openai2024embeddings} (semantic task embedding for the teacher gate), accessed through the OpenAI API under the OpenAI Terms of Service.
\item \textbf{BrowserGym}~\cite{chezelles2025browsergym} (web-agent environment interface) and the \textbf{AgentLab} \texttt{GenericAgent} implementation, released under the Apache 2.0 license.
\item \textbf{MiniWoB}~\citep{liu2018reinforcement} and \textbf{TimeWarp}~\citep{ishmam2026timewarp}, used as evaluation benchmarks under the licenses provided by their respective authors.
\end{itemize}
All artifacts are used in a manner consistent with their intended use. We do not redistribute any of the above models or datasets.

\section{Data and Privacy}
Both benchmarks consist of synthetic or curated web-interaction tasks released by their authors for research on web agents. We use them as published, without modification, and we do not collect any new user data. The tasks do not contain personally identifying information about end users.

\section{Use of AI Assistants}
We used general-purpose LLM-based assistants to help with code drafting and writing polish (e.g., grammar and phrasing). All technical content, experimental design, and results were authored, reviewed, and verified by the authors. AI assistants were not used to generate or interpret experimental results.

\section{Potential Risks}
Our method targets web-agent adaptation in controlled benchmark environments and does not involve human subjects, harmful content generation, or deployment to live web services. As with any LLM-based agent, deploying derivatives of this work to real websites would require additional safety considerations (rate limiting, action whitelisting, error monitoring) that are out of scope for this paper.

\end{document}